\documentclass[letterpaper]{article} 
\usepackage[table,xcdraw,x11names]{xcolor}
\usepackage{aaai2027}  
\usepackage[hyphens]{url}  
\usepackage{graphicx} 
\usepackage{natbib}  
\usepackage{caption} 
\usepackage{algorithm}
\usepackage{algorithmic}

\usepackage{newfloat}
\usepackage{listings}
\DeclareCaptionStyle{ruled}{labelfont=normalfont,labelsep=colon,strut=off} 
\floatstyle{ruled}
\newfloat{listing}{tb}{lst}{}
\floatname{listing}{Listing}

\usepackage{booktabs}

\usepackage{tabularx}
\usepackage{subcaption}

\title{MLPerf Automotive}
\author{
    Radoyeh Shojaei\textsuperscript{\rm 1},
    Predrag Djurdjevic\textsuperscript{\rm 2},
    Mostafa El-Khamy\textsuperscript{\rm 3},
    James Goel\textsuperscript{\rm 4},
    Kasper Mecklenburg\textsuperscript{\rm 2},\\
    P\i{}nar Muyan-\"{O}z\c{c}elik\textsuperscript{\rm 5},
    John Owens\textsuperscript{\rm 1},
    Tom St.\ John\textsuperscript{\rm 6},
    Jinho Suh\textsuperscript{\rm 7},
    Arjun Suresh\textsuperscript{\rm 8}
}
\affiliations{
    \textsuperscript{\rm 1}University of California, Davis
    \textsuperscript{\rm 2}Arm
    \textsuperscript{\rm 3}Samsung
    \textsuperscript{\rm 4}Qualcomm \\
    \textsuperscript{\rm 5}California State University, Sacramento
    \textsuperscript{\rm 6}Gilmet Labs
    \textsuperscript{\rm 7}NVIDIA
    \textsuperscript{\rm 8}AMD
}

\begin{document}

\maketitle

\begin{abstract}
We present MLPerf Automotive, the first standardized public performance benchmark for evaluating Machine Learning systems that are deployed for AI acceleration in automotive systems.
Developed through a collaborative partnership within MLCommons, this benchmark addresses the need for standardized performance evaluation methodologies in automotive machine learning systems.
Existing benchmark suites cannot be utilized for these systems since automotive workloads have unique constraints including sensor suites, safety, and real-time processing that distinguish them from the domains that previously introduced benchmarks target.
Our implemented and adopted MLPerf Automotive benchmark is a framework for evaluation and methodology for benchmarking automotive systems with reproducible performance metrics.
The benchmark consists of automotive perception tasks in 2D object detection, 2D semantic segmentation, 3D object detection, end-to-end driving, and an infotainment system application. We carefully curated and customized models for automotive use cases. We describe the methodology behind the benchmark design including the task selection, reference models, and submission rules.
We also discuss the challenges involved in acquiring the datasets and the engineering efforts to develop the reference implementations.
Our benchmark code is available at \url{https://github.com/mlcommons/mlperf_automotive}.
\end{abstract}


\section{Introduction}

Machine learning (ML) models are used in a variety of automotive applications. These applications include computer vision for Advanced Driver Assistance Systems (ADAS)~\cite{Janai:2020:CVF}, infotainment~\cite{Ju:2024:CAO}, and predictive maintenance~\cite{Jain:2022:SLR}. Within the ADAS domain, there are various levels of autonomy provided by systems. Table~\ref{tab:sae_levels} shows different levels of driving automation as defined by SAE International~\cite{SAE:2021:TAD}. The compute and safety requirements of systems increase as more advanced levels of autonomy are deployed. Automotive System-on-Chips (SoCs) available today demonstrate this wide spectrum of requirements. Theoretical peak performance varies dramatically, spanning from five teraoperations per second (TOPS)~\cite{Mobileye:2025:ETS} to one thousand TOPS~\cite{NVIDIA:2025:DAD}. Power consumption scales correspondingly with these computational demands, creating broad ranges that depend heavily on the specific application. Automotive SoCs' power usage can range from tens~\cite{Lux:2021:AIC, Mobileye:2024:MNN} to hundreds~\cite{NVIDIA:2025:DAT} of watts depending on the system configuration. These compute and power requirements are both higher than mobile devices~\cite{Gupta:2024:3WO} and lower than high-end servers~\cite{MLCommons:2025:MID, Wang:2023:PEO, Zhu:2024:APC}.

Two other key automotive requirements are lifespan and safety. Automotive hardware has longer life cycles, as the lifespan of passenger vehicles ranges from 9 to 23 years~\cite{Oguchi:2015:RAL}. For automotive chips to operate reliably and safely for a long period of time, automotive SoCs must meet rigorous mechanical stress testing~\cite{AEC:2023:AQF} and functional safety requirements~\cite{ISO:2018:RVF} that exceed those found in other computing environments.

\begin{table*}[t]
\scriptsize
\renewcommand{\arraystretch}{2.0}
\caption{SAE International levels of driving automation~\cite{SAE:2021:TAD}. Blue indicates where a human driver is actively engaged in driving. Yellow indicates an automated system is driving.}%
\label{tab:sae_levels}

\definecolor{saeblue}{HTML}{76A5AF}
\definecolor{saegold}{HTML}{BF9000}

\begin{tabularx}{\textwidth}{>{\raggedright\arraybackslash}p{0.1\textwidth} *{6}{>{\raggedright\arraybackslash}X}}
\toprule
 & \multicolumn{1}{c}{\cellcolor{saeblue}Level 0} & \multicolumn{1}{c}{\cellcolor{saeblue}Level 1} & \multicolumn{1}{c}{\cellcolor{saeblue}Level 2} & \multicolumn{1}{c}{\cellcolor{saegold}Level 3} & \multicolumn{1}{c}{\cellcolor{saegold}Level 4} & \multicolumn{1}{c}{\cellcolor{saegold}Level 5} \\

Name & \cellcolor{saeblue}No Automation & \cellcolor{saeblue}Driver Assistance & \cellcolor{saeblue}Partial Driving Automation & \cellcolor{saegold}Conditional Driving Automation & \cellcolor{saegold}High Driving Automation & \cellcolor{saegold}Full Driving Automation \\

Definition & \cellcolor{saeblue}Human driver in full control even when enhanced by active safety systems. & \cellcolor{saeblue}System performs lateral \emph{or} longitudinal vehicle motion, but not both. & \cellcolor{saeblue}System performs lateral \emph{and} longitudinal vehicle motion. & \cellcolor{saegold}System drives under limited conditions. Human driver is ready to take over if needed. & \cellcolor{saegold}System drives under limited conditions. Human driver is not expected to take over. & \cellcolor{saegold}System drives under all conditions. \\

Human Driver Engagement &
\multicolumn{3}{>{\hsize=\dimexpr 3\hsize+4\tabcolsep\relax\raggedright\arraybackslash}X}{\cellcolor{saeblue}Human driver is constantly engaged when support features are active.} &
\multicolumn{1}{>{\raggedright\arraybackslash}X}{\cellcolor{saeblue}Human driver takes over when requested.} &
\multicolumn{2}{>{\hsize=\dimexpr 2\hsize+2\tabcolsep\relax\raggedright\arraybackslash}X}{\cellcolor{saegold}Human driver is not required to take over.} \\
\bottomrule
\end{tabularx}
\end{table*}

In addition to chip design, automotive ML applications are different from other domains. ML driving workloads are much more perception-centric, requiring real-time processing of environmental data to enable autonomous decision making. As a safety-critical system, automotive ADAS applications impose stringent latency constraints that exceed those of conventional applications. Additionally, automotive models process more sensor data, whether from different modalities or multiplicities from the same modality. This sensor diversity requires models to handle features from different sensors as well as inputs that can be both dense (image arrays) and sparse (Radar, LiDAR point clouds).

\begin{figure}
  \centering
  \includegraphics[width=\linewidth]{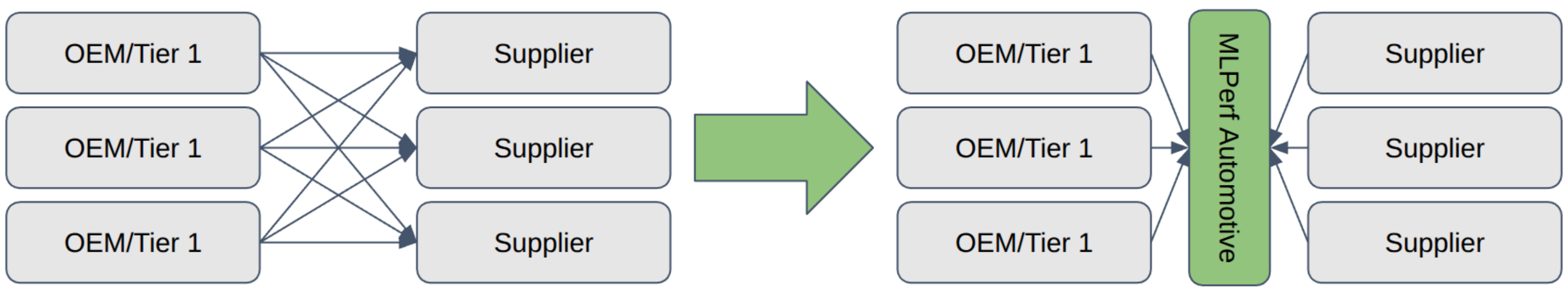}
  \caption{The goal of standardizing the benchmarking process for automotive system suppliers. On the left is the complicated individual benchmarking process and on the right is the standardized use of MLPerf Automotive.}%
  \label{fig:supplier}
\end{figure}

These requirements distinguish automotive workloads from other ML domains. Existing benchmark suites that focus on datacenter~\cite{Reddi:2020:MIB}, mobile~\cite{Reddi:2022:MMI}, or IoT~\cite{Banbury:2021:MTB} do not address automotive-specific requirements as they are not relevant in those domains. Table~\ref{tab:diff_inference} enumerates the most important differences in the scope of MLPerf Automotive vs.\ MLPerf Inference. Benchmark results in other domains cannot be used to infer performance in automotive systems because of the differences in systems and ML applications. The current process for evaluating automotive systems is performed individually by suppliers, as shown in Figure~\ref{fig:supplier}. Our goal is to introduce a benchmark to standardize and guide the evaluation of automotive systems performance.

\begin{table*}[]
    \small
    \centering
    \caption{Differences between the scopes of MLPerf Automotive and MLPerf Inference.}%
    \label{tab:diff_inference}
    \begin{tabular}{lll}
    \toprule
    & MLPerf Automotive & MLPerf Inference \\ \midrule
    Domain & Automotive & General inference with datacenter focus \\
    Tasks & Perception for driving &  Language, vision, speech, etc. \\
    Datasets & Driving scenes & Varied text, speech, and image \\ \bottomrule
    \end{tabular}

\end{table*}

In this paper, we present MLPerf Automotive, a suite of automotive inference benchmarks developed by MLCommons, led by key companies in the automotive industry, to address the need for standardized performance evaluation of representative automotive ML workloads. We establish the basis for benchmarking ML systems that accelerate AI workloads for automotive systems and present the second iteration and first major release (v1.0) of the MLPerf Automotive benchmark, which covers in-vehicle infotainment (IVI), Advanced Driver Assistance Systems (ADAS), and Autonomous Driving (AD). MLPerf Automotive provides a unified methodology for measuring inference performance across diverse hardware platforms. By establishing standardized metrics and evaluation protocols, this benchmark suite enables fair comparison of ML inference capabilities across different architectures. The benchmark addresses unique automotive constraints including real-time processing and accuracy requirements that distinguish automotive ML workloads from datacenter, IoT, or less safety-critical edge applications.

The first version of the benchmark introduces three workloads that use different AI models: (1)~2D object detection with SSD~\cite{Lio:2016:SSD}; (2)~2D semantic segmentation with DeepLabv3+~\cite{Chen:2018:EDW}; and (3)~3D object detection with BEVFormer-tiny~\cite{Li:2022:BEV}.
In our second iteration we added UniAD~\cite{Hu:2023:POA} for end-to-end driving and Llama-3.1~\cite{Grattafiori:2024:TL3} for in-vehicle infotainment (IVI\@).
These workloads are representative of automotive workloads that span ADAS, perception for AD, and IVI applications. In addition to automotive workloads, we have introduced inference rules that are applicable to safety-critical edge systems yet still accessible to submitters. The primary differences between the automotive benchmark and other inference benchmarks are the following.
\begin{itemize}
\item We select tasks, datasets, and models that are relevant to the automotive domain.
\item The benchmark performance metrics reflect the real-time safety needs in ADAS and AD systems.
\end{itemize}

The MLPerf Automotive Working Group meetings have broad attendance, covering the entire automotive supply chain, from IP and SOC suppliers, to OS vendors and the final system integrations and carmakers; Tier 1s and OEMs. There are already strong indications that benchmarks in the MLPerf Automotive benchmark suite are being adopted into the Request for Information/Quotation (RFI/RFQ) process, which is a key phase in the evaluation, and eventually, selection of automotive ML systems which go into vehicles. Because the producers and consumers of ML performance data are in the same room, the benchmark suite is ensured to contain key representative workloads, aligning the entire automotive supply chain around ML benchmarking. This is what makes MLPerf Automotive unique and valuable for the automotive industry.

\section{Benchmark Design}%
\label{benchmark}
The goal of the benchmark suite is to measure system performance on workloads across different levels of automation, from driver-assist features to full self-driving functionalities. The compute requirements vary across the spectrum of automated driving functionality, so the tasks we benchmark should be representative of these differences. We chose 2D object detection, 2D semantic segmentation, and 3D object detection for our initial benchmark and describe the reasoning in Section~\ref{benchmark:model}.

We maintain the same broad set of principles from the MLPerf Inference benchmark~\cite{Reddi:2020:MIB} that shape the key aspects of our benchmark which are summarized here. We provide a reference implementation that defines the operations. Submitters are responsible for the system under test (SUT). The dataset, Load Generator (LoadGen), and accuracy scripts are provided by MLPerf. There are two categories for submissions, closed and open. The closed division is intended to have fair comparisons between systems and have a set of associated rules. We prohibit retraining the models, caching results, and benchmark aware preprocessing.

\begin{figure}
  \centering
  \includegraphics[width=\linewidth]{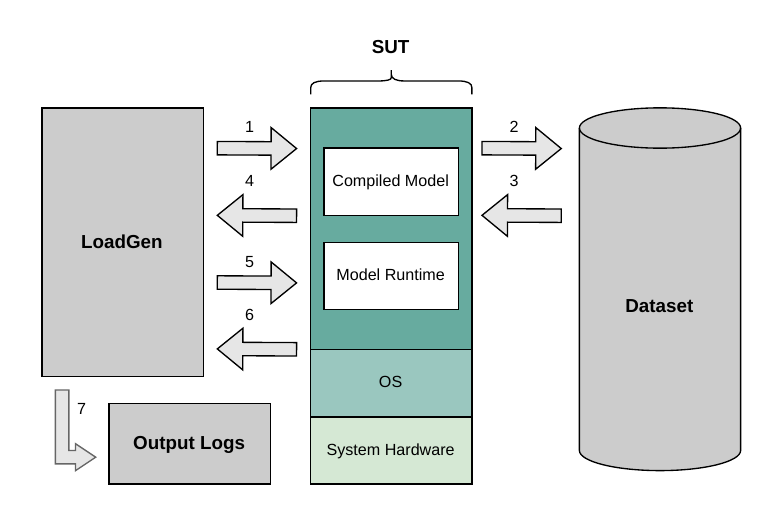}
  \caption{ A system under test (SUT) during an inference run. (1) Setup benchmark, model, dataset, pre/post processing. (2) LoadGen creates queries of Sample IDs from the dataset for SUT\@. (3) Load samples into memory. (4) SUT is ready. (5) Issue request to SUT\@. (6) SUT return results and results are post-processed. (7) Logs output for latency and accuracy analysis.}%
  \label{fig:SUT_diagram}
\end{figure}

A basic diagram of a benchmark run is shown in Figure~\ref{fig:SUT_diagram}. The SUT will be setup and load samples into memory. The LoadGen will issue requests to the SUT where results are sent back to the LoadGen for latency and accuracy assessment. Submitters submit two runs of the reference implementations. A performance run to gather performance metrics (tail latencies) and an accuracy run to verify that the model specific accuracy constraints are met. Submitters must also run compliance tests for each benchmark to ensure rules are followed. The open division relaxes closed division rules and allows retraining, different models, etc.

Table~\ref{tab:benchmarks} shows an overview of the benchmarks. In the next sections we discuss specific changes and requirements we made for the automotive benchmark.
We discuss our guiding principles for scenarios, reference implementations, and requirements we set for submissions.

\begin{table*}[ht]
\centering
\caption{Overview of the automotive benchmark suite. The tail latency and accuracy are both expressed as percentiles. The accuracy constraint is a percentage of the reference model accuracy that needs to be met. The first three models were introduced in v0.5 and the last two added after the v0.5 submissions. The reference model tail latency was measured using a V100 GPU for all models except Llama-3.1, which was done on an H100.}%
\label{tab:benchmarks}
\begin{small}
\setlength{\tabcolsep}{5.5pt} 
\begin{tabular}{l@{\hspace{2pt}}cccccccc}
\toprule
Model & \begin{tabular}[c]{@{}c@{}}Tail\\ latency (\%)\end{tabular} & \begin{tabular}[c]{@{}c@{}}Accuracy\\ constraint (\%)\end{tabular} & Params & \begin{tabular}[c]{@{}c@{}}Images/\\ query\end{tabular} & \begin{tabular}[c]{@{}c@{}}Image\\ resolution\end{tabular} & \begin{tabular}[c]{@{}c@{}}Target\\ SAE\end{tabular} & \begin{tabular}[c]{@{}c@{}}Tail\\ latency (ms)\end{tabular} \\ \midrule
BEVFormer-tiny & 99.9 & 99   & 45M & 6 & 800$\times$450   & $\geq$ 3 & 100  \\
SSD            & 99.9 & 99.9 & 14M & 1 & 3840$\times$2160 & $<$ 3    & 592  \\
DeepLabv3+     & 99.9 & 99.9 & 40M & 1 & 3840$\times$2160 & $\leq$ 3 & 1728 \\
UniAD          & 99.9 & 99   & 85M & 6 & 800$\times$450   & $\geq$ 4 & 614  \\
Llama-3.1      & 90   & 95   & 8B  & --- & ---            & ---      & 58  \\ \bottomrule
\end{tabular}
\end{small}
\end{table*}

\subsection{Scenarios}
\begin{figure}[htbp]
  \centering
  \vspace{-10pt}
  \includegraphics[width=\linewidth]{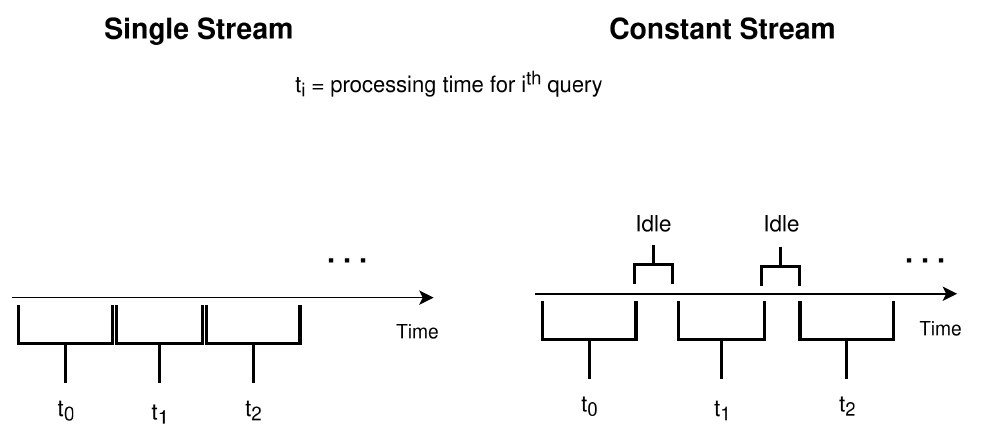}
  \caption{Benchmark scenarios}%
  \label{fig:scenarios}
\end{figure}

Automotive ML workloads can be run at different frequencies depending on the task and model complexity. A computationally expensive workload may not be able to be run at the frequency of faster sensor sampling rates and is run as fast as possible. In contrast, other workloads might only need to be run periodically. As a starting point, we introduced two scenarios that help measure peak performance as well as inference task turnaround latency under a latency deadline. One scenario is \emph{Single Stream}, which is inherited from the MLPerf Inference benchmark~\cite{Reddi:2020:MIB}. LoadGen will send a query and as soon as it is processed another query is sent, allowing the SUT to process queries as fast as possible. The other scenario is \emph{Constant Stream}, where queries are sent at a fixed frequency (e.g., 15~FPS). The SUT can be idle in between queries. Figure~\ref{fig:scenarios} shows the timing differences between the two scenarios.

\subsection{Model Selection}
\label{benchmark:model}
We choose models for our benchmark suite representative of the different levels of autonomy from Table~\ref{tab:sae_levels}.
Lower-level systems tend to use more classical/smaller models as the computations required for newer/larger models are too resource intensive and conversely slow for these systems.
Additionally, processing high resolution images is computationally demanding, as shown in the reference latency in Table~\ref{tab:benchmarks}.
Despite being smaller and older models, SSD and DeepLabv3+ operate 1--2 orders of magnitude too slow at high resolution for real-time driving.
Processing a single high resolution image for 2D tasks on a forward facing camera is a use case we want to address in the benchmark.
SSD (Single Shot MultiBox Detector)~\cite{Lio:2016:SSD} and DeepLabv3+~\cite{Chen:2018:EDW} are representative CNN architectures in computer vision. SSD is a fast single-stage object detection framework that efficiently predicts bounding boxes and class probabilities in a unified network. DeepLabv3+ is representative of classical segmentation techniques including skip connections~\cite{Long:2015:FCN}, atrous spatial pyramid pooling~\cite{Chen:2018:DSI}, and encoder-decoder structures~\cite{Ronneberger:2015:UNC}. These networks are a good representation of popular CNN networks used in perception tasks~\cite{Janai:2020:CVF}. These models are relevant to the lower SAE levels from Table~\ref{tab:sae_levels}.

To target more advanced compute systems and tasks we chose tasks involving 3D perception and end-to-end driving.
BEVFormer utilizes Bird's Eye View (BEV)~\cite{Li:2022:BEV} representations with transformer architectures for autonomous driving applications. This model addresses the challenge of multi-camera 3D perception by employing spatial and temporal self-attention mechanisms that effectively aggregate features from multiple camera viewpoints into a unified BEV space, enabling robust 3D object detection with only cameras. BEVFormer's use of deformable attention and temporal modeling allows it to capture both spatial relationships across different camera views and temporal consistency across video frames. Deformable attention enables efficient computation overhead using less compute elements but more complex elements relative to standard attention, improving the inference speed~\cite{Zhu:2021:DDD}. The mechanism for fusing camera features is also representative of multi-modal models that fuse features from different sensor modalities~\cite{Liu:2022:BMT}.
This makes BEVFormer a representative model of the use of transformer architectures in 3D computer vision~\cite{Lahoud:2022:3VW}. We chose BEVFormer for systems at SAE level 3 or higher from Table~\ref{tab:sae_levels}.
The most advanced driving task in our suite is end-to-end driving using UniAD~\cite{Hu:2023:POA}.
This is targeting SAE levels 4 and above. UniAD incorporates BEVFormer as the BEV encoder of the model and integrates a full stack of perception, prediction, and planning into a single modular network that allows safety introspection into each stage of the pipeline.

The last part of the benchmark suite is an in-vehicle infotainment (IVI) task using a large language model (LLM\@).
This is a different domain than the driving tasks involving perception, but it is an important automotive application, applicable to tasks such as navigation assistance, and one where we received significant feedback as necessary to include in the suite.
We chose Llama-3.1 8B as our reference model for the infotainment task.
MLPerf has licenses to use the Llama 3.1~\cite{Grattafiori:2024:TL3} family of LLMs so we chose among this set for the benchmark suite.
Among the 3.1 models, both the 1B and 8B parameter models are representative of model sizes used in vehicles. Our working group consensus was to choose the 8B parameter model as we expect model sizes used in IVI systems to increase over time and we wanted to be forward-looking.
\subsection{Reference Implementations}
It is important for the benchmark to provide interpretable and usable reference implementations, while achieving domain expertise. Reference implementations are provided publicly using commonly adopted frameworks. The benchmark working group members contribute reference network models in Python supporting models in either PyTorch or Open Neural Network Exchange (ONNX)~\cite{ONNX:2025:ONN}, or ideally both. All reference models can be executed on systems without accelerators and are designed in mind to easily utilize commodity AI accelerators. Having models in ONNX makes it easier for submitters to implement models on their systems. The ONNX format is a static graph which is easier for submitters to convert to their specfic intermediate representations for compilation. ONNX is also a platform-agnostic format.
This is especially important for automotive systems as the reference models need to be executable on a broad variety of hardware architectures.

The drawback of converting models to ONNX is the increased engineering effort. Public models are typically in PyTorch and don't necessarily convert easily. We consider this effort worthwhile for two reasons. First, submitters will typically convert their models from a framework like PyTorch themselves. Centralizing the engineering effort eases submitter burdens. Second, by making the engineering work easier for submitters, we aim to increase the number of submitters.
For our benchmark suite, we provide a reference implementation in either ONNX or PyTorch and in some cases both formats when within the engineering resources of our group.

\subsection{Accuracy Target}
We require submissions to maintain the reference model accuracy within some tolerance level. This is expressed as a percentage of the accuracy achieved by the reference model in the provided data set. While quantization is a standard approach for alleviating memory bandwidth bottlenecks and enhancing compute density, it frequently introduces accuracy degradation that requires sophisticated statistical or algorithmic mitigation~\cite{Rokh:2023:ACS, Lang:2024:ACS}. For closed division submissions, we permit Post-Training Quantization (PTQ) but prohibit Quantization-Aware Training (QAT), as the scope of this benchmark is strictly limited to inference performance rather than training capabilities.

Our accuracy constraints are calibrated to balance stringent automotive safety requirements with empirical feasibility given the provided dataset and benchmark rules. We use both publicly quantized models and internal experimental baselines to set the accuracy contraints. Publicly available results with BEVFormer and UniAD on NuScenes using mixed Int8 and FP16 precision achieved accuracies below 99.9\%, but above 99\%~\cite{NVIDIA:2025:DL4} of the FP32 baseline. Hence, we set 99\% as the accuracy constraint. PTQ on CNN-based models can preserve high accuracy relative to a FP32 baseline~\cite{Krishnamoorthi:2018:QDC}, which we confirmed in our internal experiments with Cognata. So, our quality targets were 99.9\% for Deeplabv3+ and SSD\@. For Llama-3.1-8B, we chose a more relaxed constraint of 95\% as this is not a safety critical task.

\subsection{Latency}
Real-time safety-critical situations have hard latency constraints, as action in emergency situations is required within millisecond timeframes.
We use tail latency as our performance metric, in a similar way to the latency-sensitive inference benchmarks used in the MLPerf Inference Edge category~\cite{Reddi:2020:MIB}. However, we need to set stronger requirements.
There is a practical trade off in that the higher the statistical significance, the longer the benchmark must be run for a valid submission because estimates of the tail latency require more samples.
We settled on a tail latency of 99.9\% performance metric. We found a 99.9\% tail latency is a good balance of capturing latency jitters without requiring excessive runtime resources.
This is stronger than previous inference benchmarks, but still allows submitters to complete a benchmark run on the order of a day rather than weeks. For Llama-3.1 8B, we chose a tail latency of 90\% as this is not a safety-critical task.

\subsection{Datasets}%
\label{benchmark:dataset}
An ideal performance benchmark dataset should contain real-world data with diverse geographic locations, weather conditions, lighting scenarios, and varying object densities within scenes. The dataset should also have a multimodal sensor suite and labeling for tasks in 2D/3D perception, planning/prediction, and end-to-end driving. There are no public datasets that meet all these requirements~\cite{Liu:2021:ASO, Liu:2024:ASO}. However, we can use a mix of datasets to meet our requirements.

The practical limitation for choosing datasets is licensing. It is common for public datasets to have non-commercial licenses. Benchmark submissions are not a direct commercial activity; however, submitters might use benchmark results as part of a customer request for quotation response. This legal ambiguity limits using any public dataset. Instead, we need to obtain explicit permission to use datasets for the benchmark.

We used three datasets for our benchmark. One is the publicly available nuScenes~\cite{Caesar:2020:NUS} dataset, which is used for BEVFormer and UniAD\@. For Llama we use the MMLU dataset~\cite{Hendrycks:2021:MMM} for general evaluation of an LLM that could be used in a vehicle dashboard. The other is a synthetic dataset obtained from Cognata~\cite{Cognata:2025:Cog} used for SSD and DeepLabv3+.

One of our requirements for a dataset was having images at 8~megapixel (MP) resolution. Typically, vision models are trained at lower resolutions, but higher-resolution images are needed for the accuracy requirements of vehicles. As of this writing, 8~MP automotive cameras are sold on the market~\cite{ZF:2025:SC6} and higher resolutions are useful for detecting small or distant objects~\cite{Ziming:2020:HHR}. Most public datasets are at lower resolutions and, combined with licensing issues, it is difficult to acquire a high-resolution image dataset.

Collecting real data is a very expensive task, so we chose to acquire a synthetic data set from Cognata~\cite{Cognata:2025:Cog}. It is common industry practice to supplement real data with synthetic to augment and diversify real data. For performance measurement, synthetic data can be used to stress-test systems in more complicated scenarios such as high-density scenes. For our benchmark suite, we want to ensure using Cognata provides useful representative performance metrics.

Table~\ref{tab:latency} shows inference performance metrics for SSD and DeepLabv3+ on the Zenseact Open Dataset (ZOD)~\cite{Alibeigi:2023:ZOD} and Cognata. Images in ZOD are real and at 8MP, matching the resolution of Cognata. For these experiments we ran our reference implementations using a single H100 GPU\@. We captured performance metrics using our benchmark suite to compare how the models performed on a real vs.\ synthetic dataset. Since our system is a shared server and not automotive-grade hardware, we expect variance in tail latencies. We examine the 50\%, 99\%, and 99.9\% tail latencies on three benchmark runs to compare the overall performance of the models. Both models had mean latencies that changed very little between the two datasets. For both SSD and DeepLabv3+, means were almost exact matches as the model's performance is determined by input resolution. For the tail latencies we see more variance, as expected. Both models showed similar low end tail latency for 99\% or 99.9\%. Overall, the model latency with Cognata has very close mean times and similar tail latency variance in our reference model tests, supporting our conclusion that performance metrics captured with Cognata are meaningfully representative of real data input.

\begin{table*}[h]
  \centering
  \small
    \caption{Latency comparison across models and datasets. Ranges are min and max values for each metric achieved across three benchmark runs.}%
    \label{tab:latency}
    \begin{tabular}{@{}llccc@{}}
        \toprule
        Model & Dataset & Mean Latency (ms) & 99\% Tail Latency (ms) & 99.9\% Tail Latency (ms) \\ \midrule
        SSD        & ZOD & 73--76 & 96--101 & 102--105 \\
        SSD        & Cognata  & 76--78 & 99--102 & 103--107 \\
        DeepLabv3+ & ZOD & 125--126 & 131--137 & 148--151 \\
        DeepLabv3+ & Cognata  & 127--129 & 133--144 & 149--155 \\ \bottomrule
    \end{tabular}
\end{table*}

\begin{figure}
  \centering
  \begin{subfigure}{.4\linewidth}
    \centering
    \includegraphics[width=\linewidth]{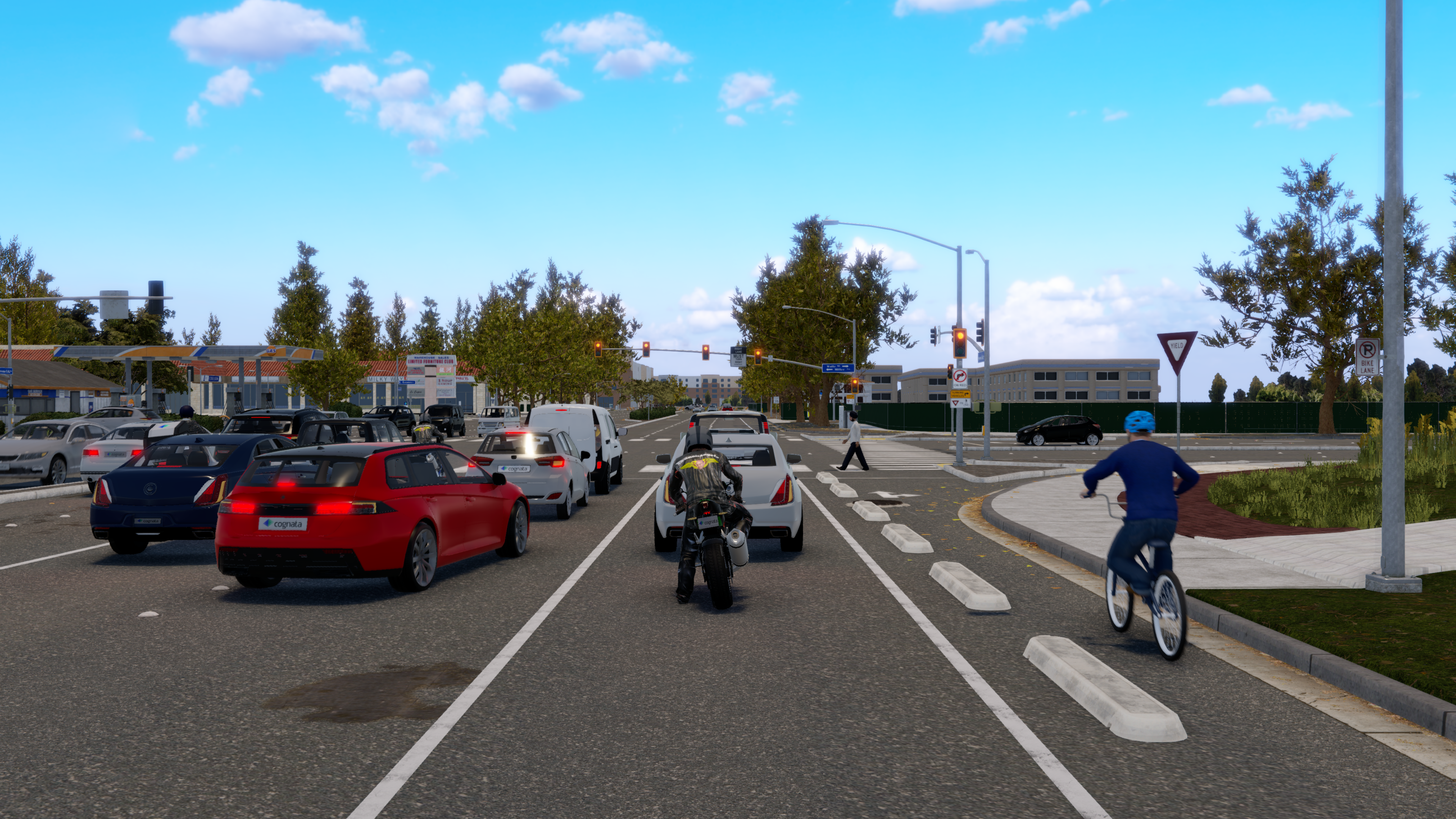}
    \caption{Cognata daytime}%
    \label{fig:cognata:urban_day}
  \end{subfigure} %
\begin{subfigure}{.4\linewidth}
    \centering
    \includegraphics[width=\linewidth]{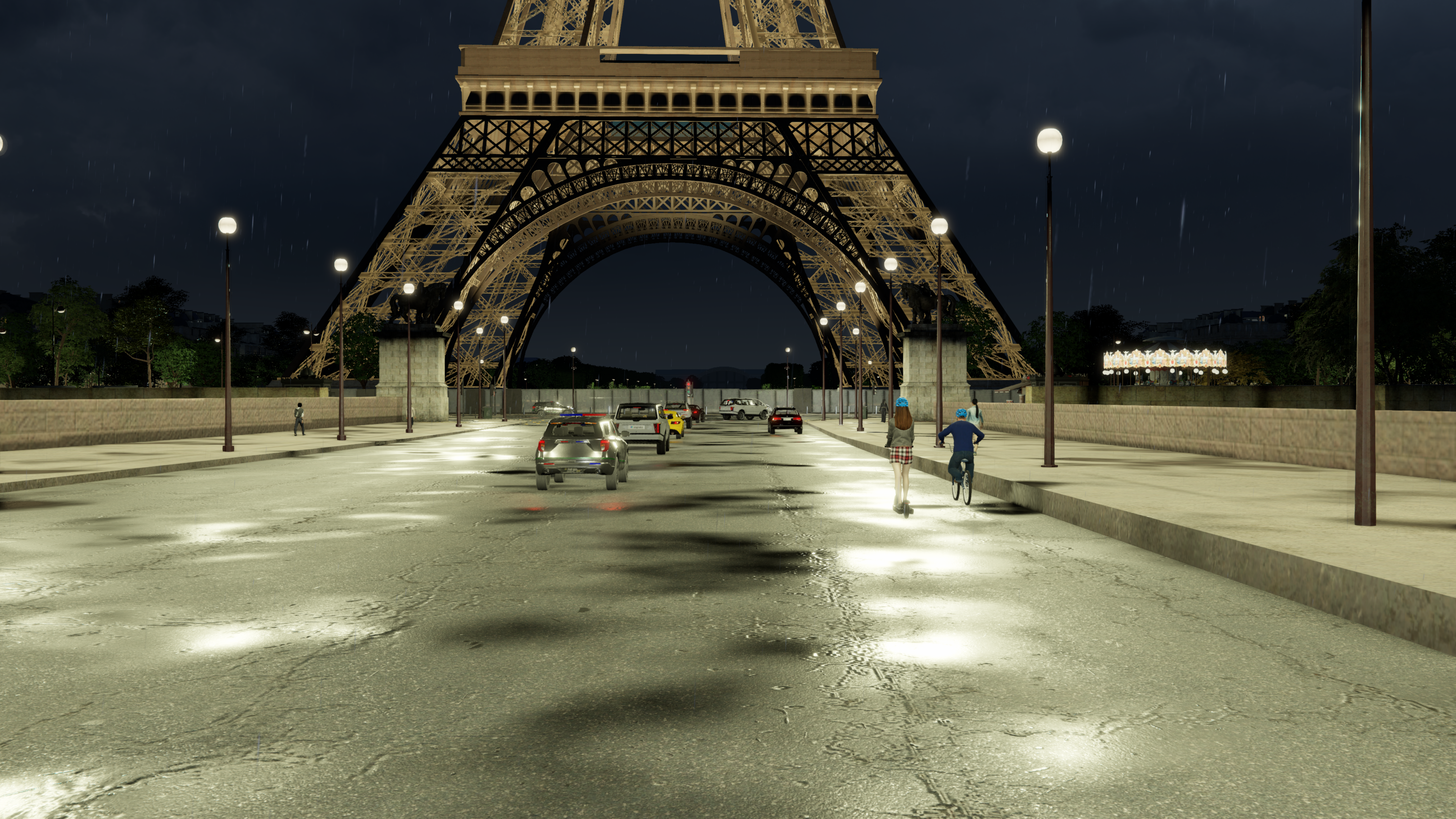}
    \caption{Cognata nighttime}%
    \label{fig:cognata:urban_night}
  \end{subfigure}
    \begin{subfigure}{.4\linewidth}
    \centering
    \includegraphics[width=\linewidth]{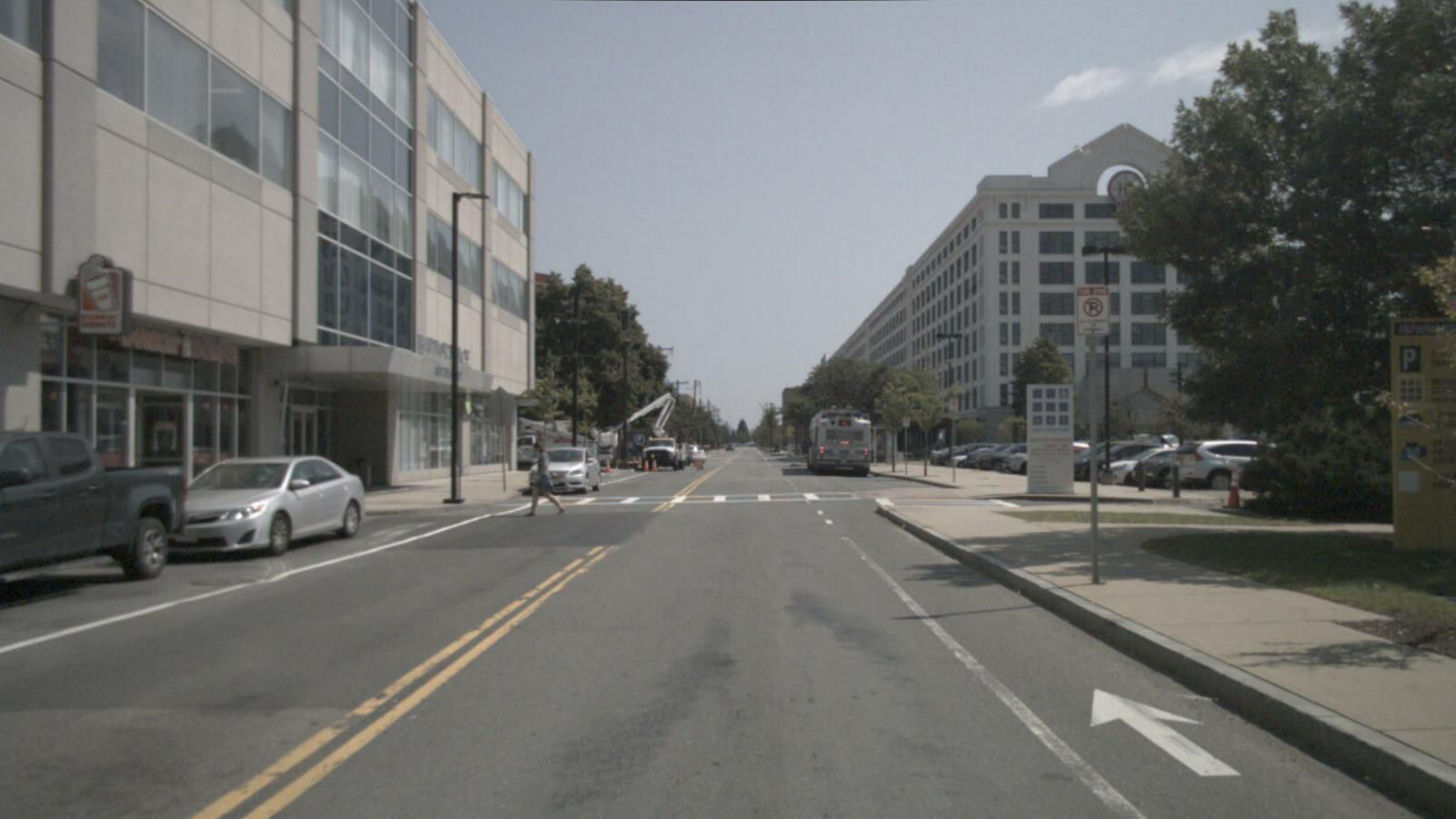}
    \caption{nuScenes daytime}%
    \label{fig:nuscenes:day}
  \end{subfigure} %
\begin{subfigure}{.4\linewidth}
    \centering
    \includegraphics[width=\linewidth]{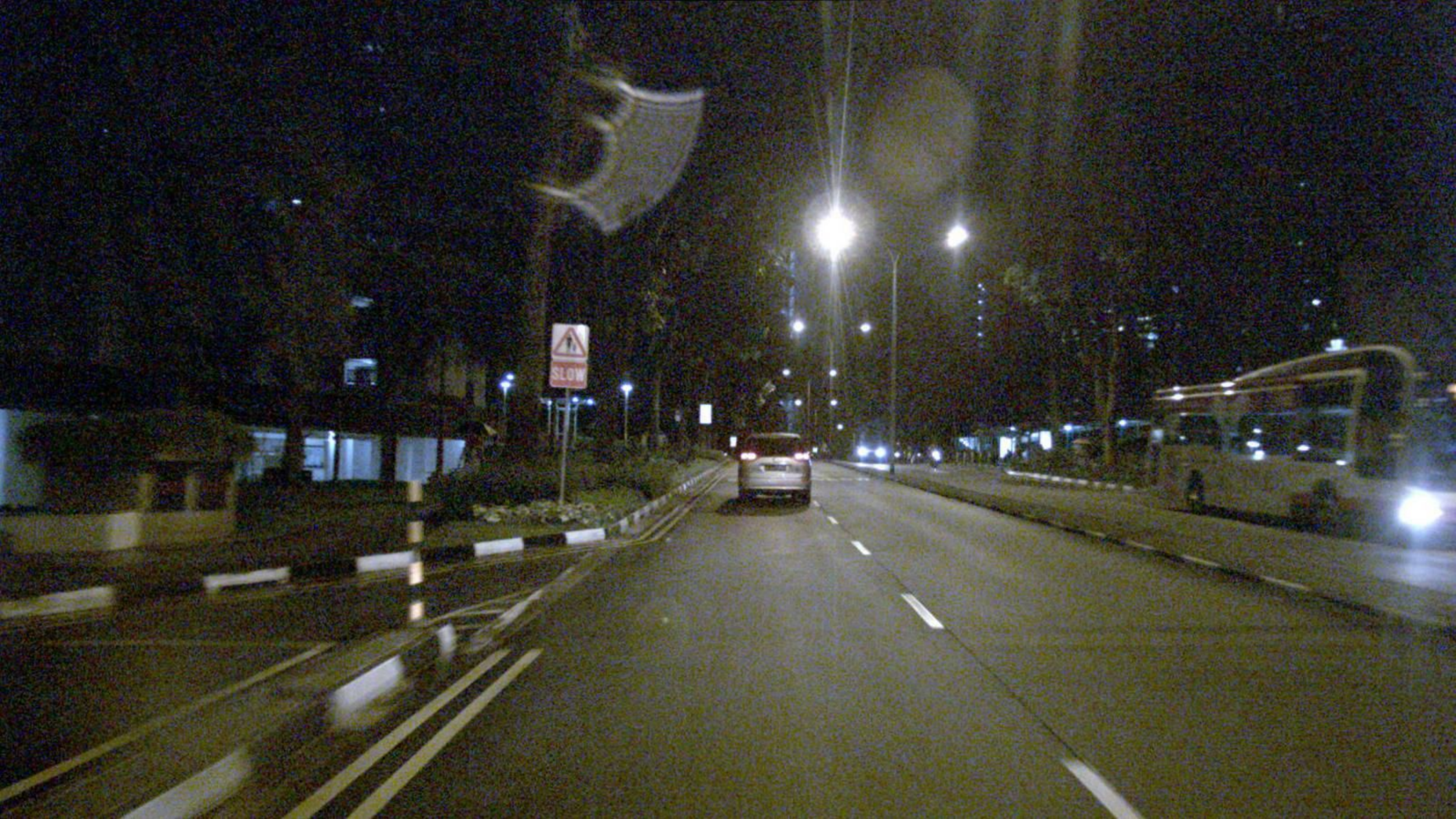}
    \caption{nuScenes nighttime}%
    \label{fig:nuscenes:days}
  \end{subfigure}
  \caption{Sample images from MLCommons Cognata dataset (top row) and nuScenes (bottom row)}%
  \label{fig:cognata}
\end{figure}

\section{Related Work}
MLPerf Automotive builds on previous inference benchmarking  and the Technical Reports on ML benchmarking recommendations for automotive from AVCC~\cite{AVCC:2021:BDN, AVCC:2023:MAD, AVCC:2024:CSF}. Other performance benchmarks are not specific to automotive applications where latency and safety are critical.

MLPerf Inference~\cite{Reddi:2020:MIB} provides a suite of benchmarks across a wide range of inference tasks. These include tasks in language, speech, and computer vision in both datacenter~\cite{MLCommons:2025:MID} and edge~\cite{MLCommons:2025:MIE} applications. Currently, the edge suite provides one automotive application~\cite{Shojaei:2025:ANA}, but otherwise the benchmarks are not relevant in an automotive context. Because the suite focuses primarily on other domains, the automotive task has received few submissions. One task is also insufficient to cover the diversity of automotive workloads. MLPerf Tiny~\cite{Banbury:2021:MTB}, Mobile~\cite{Reddi:2022:MMI}, and Client~\cite{MLCommons:2025:MC} are additional inference benchmarks specific to low-power devices, mobile devices, and personal computers. These are all edge applications with no strong real-time latency or safety requirements.

The development and evaluation of autonomous driving systems is based heavily on standardized benchmarks that provide datasets for training and testing computer vision and perception algorithms. Although these datasets benchmark model quality and not system performance, the benchmarks were important in standardizing important tasks for autonomous driving.
These datasets influenced our decisions in what content we needed when procuring the Cognata dataset as we wanted to incorporate what was useful from them into the synthetic dataset.
Additionally, the dataset challenges and results were important factors in our task and model selections for the performance benchmark. The challenge leaderboards showed what types of models were representative of specific automotive tasks.

The KITTI dataset~\cite{Geiger:2012:AWR} pioneered automotive computer vision benchmarks by providing stereo camera images, LiDAR point clouds, inertial measurement units, and GPS data collected from urban, residential, and highway scenarios around Karlsruhe, Germany, establishing evaluation protocols for tasks including stereo estimation, optical flow, visual odometry, and 3D object detection. Cityscapes~\cite{Cordts:2016:TCD} focuses specifically on semantic urban scene understanding with high-quality pixel-level annotations across 19 semantic classes for 5,000 finely annotated images and 20,000 coarsely annotated images captured in 50 cities. The nuScenes dataset~\cite{Caesar:2020:NUS}  provided a large-scale multimodal incorporating data from six cameras, five radars, and one LiDAR sensor with 3D object annotations and tracking information. The Waymo Open Dataset~\cite{Sun:2020:SIP} contains over 1,000 driving segments with high-resolution LiDAR and camera data, while providing both perception and motion forecasting benchmarks. The Mapillary Vistas~\cite{Neuhold:2017:TMV} is a globally diverse street-level imagery dataset with pixel-accurate annotations for 66 object categories across 25,000 high-resolution images spanning six continents, emphasizing geographic diversity and varied urban environments to address dataset bias issues. The ApolloScape~\cite{Wang:2020:TAO} benchmarks scene parsing, car instance identification, lane segmentation, and self-localization tasks with data captured in Chinese urban driving scenarios.

\section{Challenges and Lessons Learned}
Two major roadblocks during the benchmark development effort were: (1)~acquiring datasets and (2)~training SSD on Cognata. We refer the reader to Section~\ref{benchmark:dataset} for our decision-making process and challenges in selecting datasets for the benchmark. We also reflect on decisions we made early in development to get the wider automotive industry more involved in the benchmark development.

\begin{table}{ht}
\centering
\small
\caption{mAP of different SSD variants. Scales refers to improving anchor box scaling to match the MLCommons Cognata dataset.}%
\label{tab:ssd_map}
\begin{tabular}{ll} \toprule
Model                    & mAP \\ \midrule
BSS (baseline)           & 0.6483 \\
+scales                  & 0.6648 \\
+scales+Feature map              & 0.6943 \\
+scales+5$\times$5Detection Head     & 0.6767 \\
+scales+both & 0.7141 \\ \bottomrule
\end{tabular}
\end{table}

\begin{figure}
  \centering
  \includegraphics[width=0.8\linewidth]{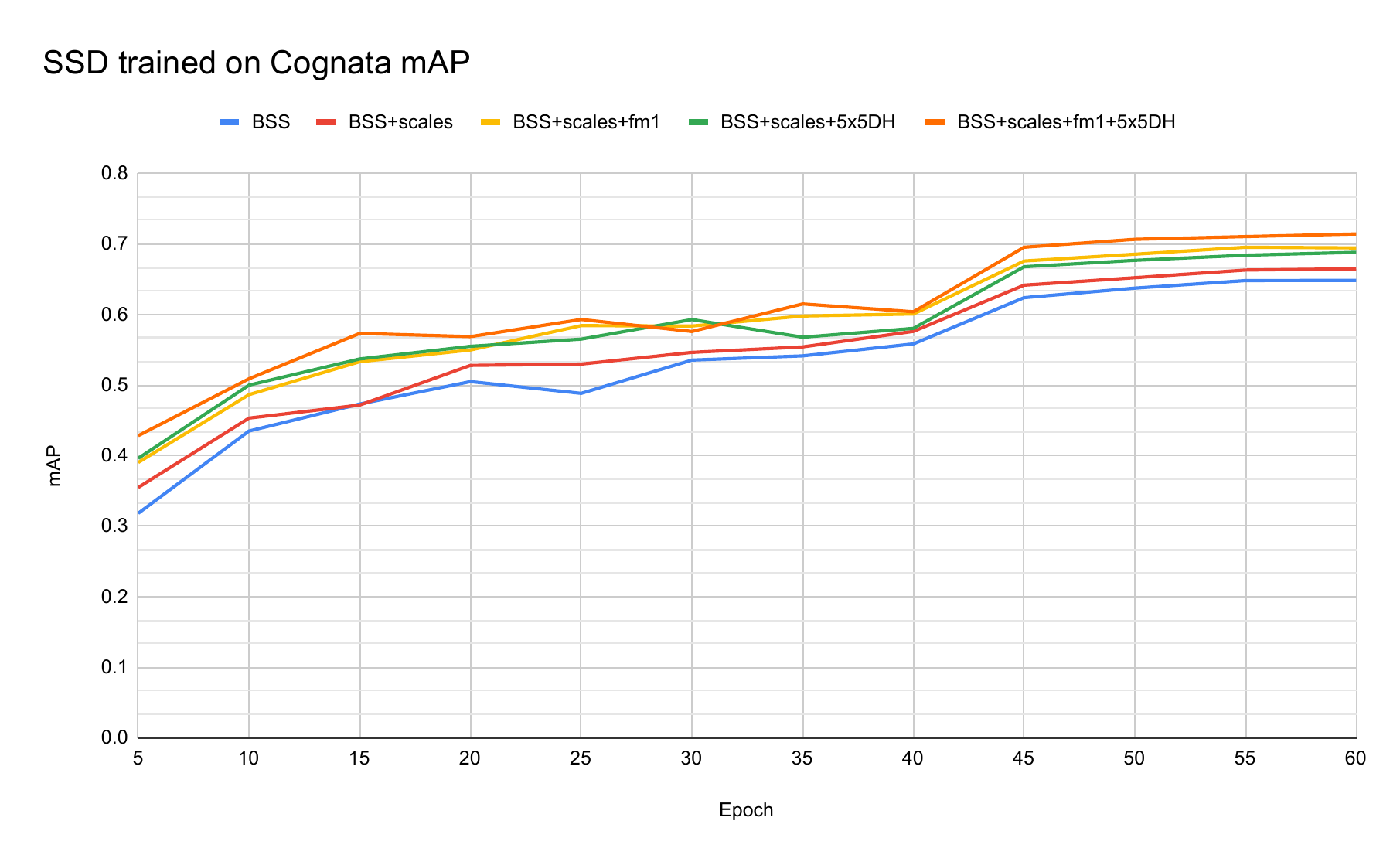}
  \caption{SSD trained on the MLCommons Cognata dataset for 60 epochs. The variant with the best accuracy showed immediate benefit in the first epoch and maintained better accuracy until accuracy plateaued for all variants.}%
  \label{fig:ssd}
\end{figure}

Training on Cognata at 8~MP required non-trivial training resources to be able to train and adjust the model.
With eight NVIDIA H100s, training SSD on Cognata for 60 epochs took about 1.5 days.
Additionally, we needed to tune the model to achieve an accuracy we considered acceptable for the benchmark.
Figure~\ref{fig:ssd} shows the accuracy results with different SSD variations. We started with a minimally changed baseline with anchor box feature sizes and steps (BSS) to match the higher resolution Cognata images.
The next step was to modify anchor box scaling (BSS+scales). The original SSD source code was used on COCO~\cite{Lin:2014:MCC}, which has lower image resolution.
Objects in our dataset have more pixels per object.
Additionally, we have both small/far objects and large/close objects so the scales for anchor boxes needed to reflect the varied object sizes in our dataset.
This increased the mean Average Precision (mAP) by 0.0165. We tested two further changes, increasing the kernel size from 3$\times$3 to 5$\times$5 and adding an additional feature map prior to the detection head (increasing from six feature maps to seven). Both changes improved the model accuracy and combining both yielded the best mAP of 0.7141 as shown in Table~\ref{tab:ssd_map}. Although it is possible to improve the model further, we do not need the best possible accuracy for a performance benchmark. We only need a model with accuracy that is representative of what deployed models can achieve given the dataset.

When initially developing the benchmark, we developed a proof of concept by implementing SSD with a demo video. The goal was to generate more interest in the benchmark, both to get more members of the automotive domain involved and to get more submissions for our first round of results. Overall, the proof of concept did not generate that much interest. It was not until we had official results that we started to gain more input and traction from the automotive industry.

\section{Future Work}

The automotive industry is moving quickly with respect to machine learning. MLPerf Automotive must update its benchmark suites to keep up with this pace. One interesting new development is an increasing shift to using multi-modal models, including vision-language-action models (VLA), for end-to-end (E2E) autonomous driving~\cite{Jian:2025:ASO}. We believe this is the future direction of automotive models and are looking to incorporate a VLA model into the benchmark, as well as classical safety oriented counterpart models to VLAs.

Beyond expanding the current model and task portfolio, we plan to introduce three significant enhancements to the evaluation framework. First, we will establish a submission category for pre-silicon system evaluations, enabling performance assessment during earlier development phases. Second, we will incorporate standardized power measurement protocols to provide comprehensive power metrics along with performance benchmarks. Third, we will incorporate more sophisticated safety-centric accuracy metrics such as difficult objects, rare objects, zero shot, and temporal accuracy clustering. These additions will strengthen MLPerf Automotive's capability to evaluate both emerging hardware architectures and the critical power constraints inherent in automotive systems.

\section{Conclusion}
This work presents MLPerf Automotive, a comprehensive benchmark suite specifically designed to evaluate the performance of machine learning systems in automotive applications. Building upon the foundation of established MLPerf inference benchmarks, we have developed a specialized framework that addresses the unique requirements and constraints of automotive computing environments.

We introduced five automotive tasks in our benchmark suite involving perception, end-to-end driving, and IVI. We defined suitable rules for safety critical real-time applications. In our ongoing development, we will update the benchmark with additional models and scenarios.

\bibliography{refs}

@techreport{AEC:2023:AQF,
  author =     {{Automotive Electronics Council}},
  title =     {{AEC-Q100}: Failure Mechanism Based Stress Test
                  Qualification for Integrated Circuits},
  institution =     {{AEC Component Technical Committee}},
  year =     2023,
  month =     aug,
  number =     {Rev-J},
  url = {http://www.aecouncil.com/Documents/AEC_Q100_Rev_J_Base_Document.pdf},
  note =     {Base Document \url{http://www.aecouncil.com/Documents/AEC_Q100_Rev_J_Base_Document.pdf}}
}

@techreport{AVCC:2021:BDN,
  author =     {AVCC},
  title =     {Benchmarking Deep Neural Network ({DNN}) for
                  Automated and Assisted Driving Systems},
  number =     {TR-003},
  year =     2021,
  institution =     {AVCC},
  url =         {https://avcc.org/tr003/},
  note = {\url{https://avcc.org/tr003/}}
}

@techreport{AVCC:2023:MAD,
  author =     {AVCC},
  title =     {Models and Datasets for Benchmarking Deep Neural
                  Networks for Automated and Assisted Driving Systems},
  number =     {TR-004},
  year =     2023,
  institution =     {AVCC},
  url =         {https://avcc.org/tr004/},
  note = {\url{https://avcc.org/tr004/}}
}

@techreport{AVCC:2024:CSF,
  author =     {AVCC},
  title =     {Compute Scenarios for Benchmarking Machine Learning
                  for Automated and Assisted Driving Systems},
  number =     {TR-007},
  year =     2024,
  institution =     {AVCC},
  url =         {https://avcc.org/tr007/},
  note = {\url{https://avcc.org/tr007/}}
}

@inproceedings{Banbury:2021:MTB,
  title =     {{MLPerf} {T}iny Benchmark},
  author =     {Banbury, Colby and Reddi, Vijay Janapa and Torelli,
                  Peter and Holleman, Jeremy and Jeffries, Nat and
                  Kiraly, Csaba and Montino, Pietro and Kanter, David
                  and Ahmed, Sebastian and Pau, Danilo and others},
  booktitle =     {Proceedings of the Neural Information Processing
                  Systems Track on Datasets and Benchmarks},
  year =     2021,
  month =     dec,
  url =         {https://openreview.net/forum?id=8RxxwAut1BI},
}

@inproceedings{Caesar:2020:NUS,
  author =     {Caesar, Holger and Bankiti, Varun and Lang, Alex H.
                  and Vora, Sourabh and Liong, Venice Erin and Xu,
                  Qiang and Krishnan, Anush and Pan, Yu and Baldan,
                  Giancarlo and Beijbom, Oscar},
  title =     {nu{S}cenes: A Multimodal Dataset for Autonomous
                  Driving},
  booktitle =     {Proceedings of the IEEE/CVF Conference on Computer
                  Vision and Pattern Recognition (CVPR)},
  month =     jun,
  year =     2020,
  pages =     {11618--11628},
  doi =         {10.1109/CVPR42600.2020.01164},
}

@article{Chen:2018:DSI,
  author =     {Chen, Liang-Chieh and Papandreou, George and
                  Kokkinos, Iasonas and Murphy, Kevin and Yuille, Alan
                  L.},
  journal =     {IEEE Transactions on Pattern Analysis and Machine
                  Intelligence},
  title =     {{DeepLab}: Semantic Image Segmentation with Deep
                  Convolutional Nets, Atrous Convolution, and Fully
                  Connected {CRF}s},
  year =     2018,
  volume =     40,
  number =     4,
  pages =     {834--848},
  doi =         {10.1109/TPAMI.2017.2699184}
}

@inproceedings{Chen:2018:EDW,
  title =     {Encoder-decoder with atrous separable convolution
                  for semantic image segmentation},
  author =     {Chen, Liang-Chieh and Zhu, Yukun and Papandreou,
                  George and Schroff, Florian and Adam, Hartwig},
  booktitle =     {Proceedings of the European Conference on Computer
                  Vision (ECCV)},
  pages =     {833--851},
  year =     2018,
  month =     sep,
  doi =         {10.1007/978-3-030-01234-2_49},
}

@misc{Cognata:2025:Cog,
  author =     {Cognata},
  title =     {Cognata: Supervised Gen{AI} Simulation Platform},
  url =         {https://www.cognata.com/},
  urldate =     {2025-08-25},
  howpublished = {\url{https://www.cognata.com/}},
  note = {Accessed: 2025-08-25},
  year =     2025
}

@inproceedings{Cordts:2016:TCD,
  author =     {Cordts, Marius and Omran, Mohamed and Ramos,
                  Sebastian and Rehfeld, Timo and Enzweiler, Markus
                  and Benenson, Rodrigo and Franke, Uwe and Roth,
                  Stefan and Schiele, Bernt},
  booktitle =     {2016 IEEE Conference on Computer Vision and Pattern
                  Recognition (CVPR)},
  title =     {The {C}ityscapes Dataset for Semantic Urban Scene
                  Understanding},
  year =     2016,
  pages =     {3213--3223},
  doi =         {10.1109/CVPR.2016.350}
}

@inproceedings{Geiger:2012:AWR,
  author =     {Geiger, Andreas and Lenz, Philip and Urtasun,
                  Raquel},
  booktitle =     {Proceedings of the IEEE Computer Society Conference
                  on Computer Vision and Pattern Recognition},
  doi =         {10.1109/CVPR.2012.6248074},
  isbn =     9781467312264,
  issn =     10636919,
  pages =     {3354--3361},
  title =     {Are we ready for Autonomous Driving? {T}he {KITTI}
                  Vision Benchmark Suite},
  year =     2012,
  month =     jun,
  series =     {CVPR 2012}
}

@misc{Grattafiori:2024:TL3,
  title =     {The {L}lama 3 Herd of Models},
  author =     {Aaron Grattafiori and Abhimanyu Dubey and Abhinav
                  Jauhri and Abhinav Pandey and Abhishek Kadian and
                  Ahmad Al-Dahle and Aiesha Letman and Akhil Mathur
                  and Alan Schelten and Alex Vaughan and others},
  year =     2024,
  month =     jul,
  eprint =     {2407.21783},
  archivePrefix ={arXiv},
  primaryClass = {cs.AI},
  journal =     {CoRR},
}

@inproceedings{Gupta:2024:3WO,
  author =     {Gupta, Agrim and Heidari, Adel and Kalipattapu,
                  Avyakta and Jain, Ish Kumar and Bharadia, Dinesh},
  title =     {3 {W}'s of smartphone power consumption: Who, Where
                  and How much is draining my battery?},
  year =     2024,
  isbn =     9798400704895,
  publisher_ =     {Association for Computing Machinery},
  address_ =     {New York, NY, USA},
  doi =         {10.1145/3636534.3695905},
  booktitle =     {Proceedings of the 30th Annual International
                  Conference on Mobile Computing and Networking},
  pages =     {2248--2250},
  numpages =     3,
  location =     {Washington D.C., DC, USA},
  series =     {ACM MobiCom '24},
}

@inproceedings{Hendrycks:2021:MMM,
  title =     {Measuring Massive Multitask Language Understanding},
  author =     {Dan Hendrycks and Collin Burns and Steven Basart and
                  Andy Zou and Mantas Mazeika and Dawn Song and Jacob
                  Steinhardt},
  booktitle =     {International Conference on Learning
                  Representations},
  year =     2021,
  url =         {https://openreview.net/forum?id=d7KBjmI3GmQ},
  month =     jan,
}

@inproceedings{Hu:2023:POA,
  author =     {Hu, Yihan and Yang, Jiazhi and Chen, Li and Li, Keyu
                  and Sima, Chonghao and Zhu, Xizhou and Chai, Siqi
                  and Du, Senyao and Lin, Tianwei and Wang, Wenhai and
                  Lu, Lewei and Jia, Xiaosong and Liu, Qiang and Dai,
                  Jifeng and Qiao, Yu and Li, Hongyang},
  title =     {Planning-Oriented Autonomous Driving},
  booktitle =     {Proceedings of the IEEE/CVF Conference on Computer
                  Vision and Pattern Recognition (CVPR)},
  month =     jun,
  year =     2023,
  pages =     {17853--17862},
  doi =         {10.1109/CVPR52729.2023.01712},
}

@misc{ISO:2018:RVF,
  author =     {{ISO}},
  year =     2018,
  title =     {{Road vehicles -- Functional safety}},
  number =     {{ISO 26262:2018}},
  publisher =     {{ISO, Geneva, Switzerland}},
  url =         {https://www.iso.org/publication/PUB200262.html},
  howpublished = {\url{https://www.iso.org/publication/PUB200262.html}}
}

@article{Jain:2022:SLR,
  author =     {Jain, Muskan and Vasdev, Dipit and Pal, Kunal and
                  Sharma, Vishal},
  title =     {Systematic literature review on predictive
                  maintenance of vehicles and diagnosis of vehicle's
                  health using machine learning techniques},
  journal =     {Computational Intelligence},
  volume =     38,
  number =     6,
  pages =     {1990-2008},
  doi =         {https://doi.org/10.1111/coin.12553},
  year =     2022
}

@article{Janai:2020:CVF,
  author =     {Janai, Joel and G\"{u}ney, Fatma and Behl, Aseem and
                  Geiger, Andreas},
  title =     {Computer Vision for Autonomous Vehicles: Problems,
                  Datasets and State of the Art},
  year =     2020,
  issue_date =     {Jul 2020},
  publisher =     {Now Publishers Inc.},
  address =     {Hanover, MA, USA},
  volume =     12,
  number =     {1--3},
  issn =     {1572-2740},
  doi =         {10.1561/0600000079},
  journal =     {Found. Trends. Comput. Graph. Vis.},
  month =     jul,
  pages =     {1--308},
  numpages =     308
}

@misc{Jian:2025:ASO,
  title =     {A Survey on Vision-Language-Action Models for
                  Autonomous Driving},
  author =     {Sicong Jiang and Zilin Huang and Kangan Qian and
                  Ziang Luo and Tianze Zhu and Yang Zhong and Yihong
                  Tang and Menglin Kong and Yunlong Wang and Siwen
                  Jiao and Hao Ye and Zihao Sheng and Xin Zhao and
                  Tuopu Wen and Zheng Fu and Sikai Chen and Kun Jiang
                  and Diange Yang and Seongjin Choi and Lijun Sun},
  year =     2025,
  month =     jun,
  journal =     {CoRR},
  eprint =     {2506.24044},
  archivePrefix ={arXiv},
  primaryClass = {cs.CV},
}

@misc{Krishnamoorthi:2018:QDC,
  author =	 {Raghuraman Krishnamoorthi},
  title =	 {Quantizing deep convolutional networks for efficient
                  inference: {A} whitepaper},
  journal =	 {CoRR},
  year =	 2018,
  month =	 jun,
  journal =     {CoRR},
  url =		 {http://arxiv.org/abs/1806.08342},
  eprinttype =	 {arXiv},
  eprint =	 {1806.08342},
  archivePrefix ={arXiv},
  arxivId =	 {1806.08342},
  primaryClass = {cs.CV},
}

@inproceedings{Ju:2024:CAO,
  author =     {Ju, Jangkyu and Lee, Eunju and Park, SangJun},
  title =     {Comparative Analysis of Ensemble Machine Learning
                  Models for Personalized In-Vehicle Infotainment
                  Recommendation Systems},
  booktitle =     {Adjunct Proceedings of the 16th International
                  Conference on Automotive User Interfaces and
                  Interactive Vehicular Applications},
  year =     2024,
  isbn =     9798400705205,
  publisher_ =     {Association for Computing Machinery},
  address_ =     {New York, NY, USA},
  doi =         {10.1145/3641308.3685021},
  pages =     {45--50},
  numpages =     6,
  location =     {Stanford, CA, USA},
  series =     {AutomotiveUI '24 Adjunct}
}

@misc{Lahoud:2022:3VW,
  title =     {3{D} vision with transformers: A survey},
  author =     {Lahoud, Jean and Cao, Jiale and Khan, Fahad Shahbaz
                  and Cholakkal, Hisham and Anwer, Rao Muhammad and
                  Khan, Salman and Yang, Ming-Hsuan},
  journal =     {CoRR},
  eprint =     {2208.04309},
  archivePrefix ={arXiv},
  primaryClass = {cs.CV},
  year =     2022,
  month =     aug,
}

@inproceedings{Lang:2024:ACS,
  author =     {Lang, Jiedong and Guo, Zhehao and Huang, Shuyu},
  booktitle =     {2024 4th International Conference on Artificial
                  Intelligence, Robotics, and Communication (ICAIRC)},
  title =     {A Comprehensive Study on Quantization Techniques for
                  Large Language Models},
  year =     2024,
  pages =     {224-231},
  doi =         {10.1109/ICAIRC64177.2024.10899941}
}

@inproceedings{Li:2022:BEV,
  title =     {{BEVF}ormer: Learning Bird's-Eye-View Representation
                  from Multi-Camera Images via Spatiotemporal
                  Transformers},
  author =     {Li, Zhiqi and Wang, Wenhai and Li, Hongyang and Xie,
                  Enze and Sima, Chonghao and Lu, Tong and Yu, Qiao
                  and Dai, Jifeng},
  booktitle =     {Proceedings of the European Conference on Computer
                  Vision (ECCV)},
  pages =     {1--18},
  year =     2022,
  month =     oct,
  doi =         {10.1007/978-3-031-20077-9_1},
}

@inproceedings{Lin:2014:MCC,
  title =     {Microsoft {COCO}: Common Objects in Context},
  author =     {Lin, Tsung-Yi and Maire, Michael and Belongie, Serge
                  and Hays, James and Perona, Pietro and Ramanan, Deva
                  and Doll{\'a}r, Piotr and Zitnick, C. Lawrence},
  booktitle =     {Proceedings of the European Conference on Computer
                  Vision (ECCV)},
  pages =     {740--755},
  year =     2014,
  month =     sep,
  doi =         {10.1007/978-3-319-10602-1_48},
}

@inproceedings{Lio:2016:SSD,
  title =     {{SSD}: Single Shot MultiBox Detector},
  author =     {Liu, Wei and Anguelov, Dragomir and Erhan, Dumitru
                  and Szegedy, Christian and Reed, Scott and Fu,
                  Cheng-Yang and Berg, Alexander C.},
  booktitle =     {Proceedings of the European Conference on Computer
                  Vision (ECCV)},
  pages =     {21--37},
  year =     2016,
  month =     oct,
  doi =         {10.1007/978-3-319-46448-0_2},
}

@inproceedings{Liu:2021:ASO,
  author =     {Liu, Weiyu and Dong, Qian and Wang, Pengqi and Yang,
                  Guang and Meng, Lingzhong and Song, You and Shi,
                  Yuan and Xue, Yunzhi},
  booktitle =     {2021 8th International Conference on Dependable
                  Systems and Their Applications (DSA)},
  title =     {A Survey on Autonomous Driving Datasets},
  year =     2021,
  pages =     {399--407},
  doi =         {10.1109/DSA52907.2021.00060},
}

@inproceedings{Liu:2022:BMT,
  title =     {{BEVFusion}: Multi-Task Multi-Sensor Fusion with
                  Unified Bird's-Eye View Representation},
  author =     {Liu, Zhijian and Tang, Haotian and Amini, Alexander
                  and Yang, Xinyu and Mao, Huizi and Rus, Daniela and
                  Han, Song},
  booktitle =     {IEEE International Conference on Robotics and
                  Automation (ICRA)},
  year =     2023,
  pages =     {2774--2781},
  doi =         {10.1109/ICRA48891.2023.10160968}
}

@article{Liu:2024:ASO,
  author =     {Liu, Mingyu and Yurtsever, Ekim and Fossaert,
                  Jonathan and Zhou, Xingcheng and Zimmer, Walter and
                  Cui, Yuning and Zagar, Bare Luka and Knoll, Alois
                  C.},
  journal =     {IEEE Transactions on Intelligent Vehicles},
  title =     {A Survey on Autonomous Driving Datasets: Statistics,
                  Annotation Quality, and a Future Outlook},
  year =     2024,
  volume =     9,
  number =     11,
  pages =     {7138--7164},
  doi =         {10.1109/TIV.2024.3394735},
}

@inproceedings{Long:2015:FCN,
  author =     {Long, Jonathan and Shelhamer, Evan and Darrell,
                  Trevor},
  title =     {Fully Convolutional Networks for Semantic
                  Segmentation},
  booktitle =     {Proceedings of the IEEE Conference on Computer
                  Vision and Pattern Recognition (CVPR)},
  month =     jun,
  year =     2015,
  pages =     {3431--3440},
  doi =         {10.1109/CVPR.2015.7298965},
}

@misc{Lux:2021:AIC,
  author =     {{Lux Research}},
  title =     {{AI} Chips in the Autonomous Vehicle Space},
  year =     2021,
  month =     may,
  url =
                  {https://luxresearchinc.com/blog/ai-chips-in-the-autonomous-vehicle-space/#:~:text=Mobileye%20claims%20that%20the%20most,which%20uses%20two%20EyeQ5%20chips.},
  urldate =     {2025-09-17},
  howpublished = {\url{https://luxresearchinc.com/blog/ai-chips-in-the-autonomous-vehicle-space/#:~:text=Mobileye%20claims%20that%20the%20most,which%20uses%20two%20EyeQ5%20chips.}},
  note = {Accessed: 2025-09-17}
}

@misc{MLCommons:2025:MC,
  author =     {MLCommons},
  title =     {{MLP}erf Client},
  url =         {https://mlcommons.org/benchmarks/client/},
  urldate =     {2025-08-25},
  howpublished = {\url{https://mlcommons.org/benchmarks/client/}},
  note = {Accessed: 2025-08-25},
  year =     2025
}

@misc{MLCommons:2025:MID,
  author =     {MLCommons},
  title =     {{MLP}erf Inference: Datacenter},
  url =
                  {https://mlcommons.org/benchmarks/inference-datacenter/},
  urldate =     {2025-08-25},
  howpublished = {\url{https://mlcommons.org/benchmarks/inference-datacenter/}},
  note = {Accessed: 2025-08-25},
  year =     2025,
}

@misc{MLCommons:2025:MIE,
  author =     {MLCommons},
  title =     {{MLP}erf Inference: Edge},
  url =         {https://mlcommons.org/benchmarks/inference-edge/},
  urldate =     {2025-08-25},
  howpublished = {\url{https://mlcommons.org/benchmarks/inference-edge/}},
  note = {Accessed: 2025-08-25},
  year =     2025
}

@misc{Mobileye:2024:MNN,
  author =     {Mobileye},
  title =     {Mobileye: {N}ow. {N}ext. {B}eyond. {CES} 2024 Press
                  Conference with {P}rof.\ {A}mnon {S}hashua},
  url =         {https://www.youtube.com/watch?v=uco1z54FAdA&t=2440s},
  urldate =     {2025-09-04},
  howpublished = {\url{https://www.youtube.com/watch?v=uco1z54FAdA&t=2440s}},
  note = {Accessed: 2025-09-04},
  year =     2024
}

@misc{Mobileye:2025:ETS,
  author =     {Mobileye},
  title =     {Eye{Q}: The System-on-Chip for Automotive
                  Applications},
  url =         {https://www.mobileye.com/technology/eyeq-chip/},
  urldate =     {2025-09-04},
  howpublished = {\url{https://www.mobileye.com/technology/eyeq-chip/}},
  note = {Accessed: 2025-09-04},
  year =     2025
}

@misc{NVIDIA:2025:DAD,
  author =     {NVIDIA},
  title =     {Drive {AGX} Developer Kits},
  url =         {https://developer.nvidia.com/drive/agx},
  urldate =     {2025-09-04},
  howpublished = {\url{https://developer.nvidia.com/drive/agx}},
  note = {Accessed: 2025-09-04},
  year =     2025
}

@misc{NVIDIA:2025:DAT,
  author =     {NVIDIA},
  title =     {Drive {AGX} {T}hor Development Platform},
  year =     2025,
  month =     jun,
  url =
                  {https://developer.download.nvidia.com/drive/docs/nvidia-drive-agx-thor-platform-for-developers.pdf},
  urldate =     {2025-09-17},
  howpublished = {\url{https://developer.download.nvidia.com/drive/docs/nvidia-drive-agx-thor-platform-for-developers.pdf}},
  note = {Accessed: 2025-09-17}
}

@misc{NVIDIA:2025:DL4,
  author =     {NVIDIA},
  title =     {{DL4AGX}},
  url =
                  {https://github.com/NVIDIA/DL4AGX/},
  urldate =     {2025-08-25},
  howpublished = {\url{https://github.com/NVIDIA/DL4AGX/}},
  note = {Accessed: 2025-08-25},
  year =     2025
}

@inproceedings{Neuhold:2017:TMV,
  author =     {Neuhold, Gerhard and Ollmann, Tobias and Bulo,
                  Samuel Rota and Kontschieder, Peter},
  booktitle =     {Proceedings of the IEEE International Conference on
                  Computer Vision},
  doi =         {10.1109/ICCV.2017.534},
  isbn =     9781538610329,
  issn =     15505499,
  pages =     {5000--5009},
  title =     {The {M}apillary {V}istas Dataset for Semantic
                  Understanding of Street Scenes},
  year =     2017,
  month =     oct
}

@misc{ONNX:2025:ONN,
  author =     {ONNX},
  title =     {Open Neural Network Exchange ({ONNX})},
  url =         {https://github.com/onnx/onnx},
  urldate =     {2025-09-18},
  howpublished = {\url{https://github.com/onnx/onnx}},
  note = {Accessed: 2025-09-18},
  year =     2025
}

@article{Oguchi:2015:RAL,
  author =     {Oguchi, Masahiro and Fuse, Masaaki},
  title =     {Regional and Longitudinal Estimation of Product
                  Lifespan Distribution: A Case Study for Automobiles
                  and a Simplified Estimation Method},
  journal =     {Environmental Science \& Technology},
  volume =     49,
  number =     3,
  pages =     {1738--1743},
  year =     2015,
  doi =         {10.1021/es505245q},
  note =     {PMID: 25549538}
}

@INPROCEEDINGS{Reddi:2020:MIB,
  author =     {Reddi, Vijay Janapa and Cheng, Christine and Kanter,
                  David and Mattson, Peter and Schmuelling, Guenther
                  and Wu, Carole-Jean and Anderson, Brian and Breughe,
                  Maximilien and Charlebois, Mark and others},
  booktitle =     {2020 ACM/IEEE 47th Annual International Symposium on
                  Computer Architecture (ISCA)},
  title =     {{MLP}erf Inference Benchmark},
  year =     2020,
  pages =     {446--459},
  doi =         {10.1109/ISCA45697.2020.00045}
}

@inproceedings{Reddi:2022:MMI,
  author =     {Reddi, Vijay Janapa and Kanter, David and Mattson,
                  Peter and Duke, Jared and Nguyen, Thai and Chukka,
                  Ramesh and Shiring, Ken and Tan, Koan-Sin and
                  Charlebois, Mark and others},
  booktitle =     {Proceedings of Machine Learning and Systems},
  editor =     {D. Marculescu and Y. Chi and C. Wu},
  pages =     {352--369},
  title =     {{MLPerf} Mobile Inference Benchmark: An
                  Industry-Standard Open-Source Machine Learning
                  Benchmark for On-Device {AI}},
  url =
                  {https://proceedings.mlsys.org/paper_files/paper/2022/file/a2b2702ea7e682c5ea2c20e8f71efb0c-Paper.pdf},
  volume =     4,
  year =     2022
}

@article{Rokh:2023:ACS,
  author =     {Rokh, Babak and Azarpeyvand, Ali and Khanteymoori,
                  Alireza},
  title =     {A Comprehensive Survey on Model Quantization for
                  Deep Neural Networks in Image Classification},
  year =     2023,
  issue_date =     {December 2023},
  publisher_ =     {Association for Computing Machinery},
  address_ =     {New York, NY, USA},
  volume =     14,
  number =     6,
  issn =     {2157-6904},
  doi =         {10.1145/3623402},
  journal =     {ACM Trans. Intell. Syst. Technol.},
  month =     nov,
  articleno =     97,
  pages =     {1--50}
}

@inproceedings{Ronneberger:2015:UNC,
  author =     {Ronneberger, Olaf and Fischer, Philipp and Brox,
                  Thomas},
  editor =     {Navab, Nassir and Hornegger, Joachim and Wells,
                  William M. and Frangi, Alejandro F.},
  title =     {U-Net: Convolutional Networks for Biomedical Image
                  Segmentation},
  booktitle =     {Medical Image Computing and Computer-Assisted
                  Intervention -- MICCAI 2015},
  year =     2015,
  publisher =     {Springer International Publishing},
  pages =     {234--241},
  isbn =     {978-3-319-24574-4},
  doi =         {10.1007/978-3-319-24574-4_28}
}

@techreport{SAE:2021:TAD,
  author =     {{SAE International}},
  title =     {Taxonomy and Definitions for Terms Related to
                  Driving Automation Systems for On-Road Motor
                  Vehicles},
  number =     {J3016\_202104},
  year =     2021,
  institution =     {{SAE} International},
  url =         {https://www.sae.org/standards/content/j3016_202104/},
  note = {\url{https://www.sae.org/standards/content/j3016_202104/}}
}

@misc{Shojaei:2025:ANA,
  author =     {Radoyeh Shojaei and Victor Bittorf and Predrag
                  Djurdjevic and Kasper Mecklenburg and Pınar
                  Muyan-Özçelik and John Owens and Tom {St.\ John} and
                  Jinho Suh },
  title =     {A New Automotive Benchmark for {MLP}erf Inference
                  v5.0},
  url =         {https://mlcommons.org/2025/04/auto-inference-v5/},
  urldate =     {2025-09-26},
  howpublished = {\url{https://mlcommons.org/2025/04/auto-inference-v5/}},
  note = {Accessed: 2025-09-26},
  year =     2025
}

@inproceedings{Sun:2020:SIP,
  author =     {Sun, Pei and Kretzschmar, Henrik and Dotiwalla,
                  Xerxes and Chouard, Aurelien and Patnaik, Vijaysai
                  and Tsui, Paul and Guo, James and Zhou, Yin and
                  Chai, Yuning and Caine, Benjamin and Vasudevan,
                  Vijay and Han, Wei and Ngiam, Jiquan and Zhao, Hang
                  and Timofeev, Aleksei and Ettinger, Scott and
                  Krivokon, Maxim and Gao, Amy and Joshi, Aditya and
                  Zhang, Yu and Shlens, Jonathon and Chen, Zhifeng and
                  Anguelov, Dragomir},
  title =     {Scalability in Perception for Autonomous Driving:
                  {W}aymo Open Dataset},
  booktitle =     {Proceedings of the IEEE/CVF Conference on Computer
                  Vision and Pattern Recognition (CVPR)},
  month =     jun,
  year =     2020,
  doi =         {10.1109/CVPR42600.2020.00252},
}

@article{Wang:2020:TAO,
  author =     {Wang, Peng and Huang, Xinyu and Cheng, Xinjing and
                  Zhou, Dingfu and Geng, Qichuan and Yang, Ruigang},
  doi =         {10.1109/tpami.2019.2926463},
  issn =     {0162-8828},
  journal =     {IEEE Transactions on Pattern Analysis and Machine
                  Intelligence},
  month =     oct,
  number =     10,
  pages =     {2702--2719},
  title =     {The {A}pollo{S}cape Open Dataset for Autonomous
                  Driving and its Application},
  volume =     42,
  year =     2020
}

@article{Wang:2023:PEO,
  author =     {Wang, Yewan and N{\"o}rtersh{\"a}user, David and Le
                  Masson, St{\'e}phane and Menaud, Jean-Marc},
  title =     {Potential effects on server power metering and
                  modeling},
  journal =     {Wireless Networks},
  year =     2023,
  month =     apr,
  volume =     29,
  number =     3,
  pages =     {1077--1084},
  issn =     {1572-8196},
  doi =         {10.1007/s11276-018-1882-1},
}

@misc{ZF:2025:SC6,
  author =     {ZF},
  title =     {Smart Camera 6 ({ME})},
  url = {https://www.zf.com/products/en/cars/products_77249.html},
  urldate =     {2025-09-11},
  howpublished = {\url{https://www.zf.com/products/en/cars/products_77249.html}},
  note = {Accessed: 2025-09-11},
  year =     2025
}

@inproceedings{Zhu:2021:DDD,
  title =     {Deformable {DETR}: Deformable Transformers for
                  End-to-End Object Detection},
  author =     {Zhu, Xizhou and Su, Weijie and Lu, Lewei and Li, Bin
                  and Wang, Xiaogang and Dai, Jifeng},
  booktitle =     {International Conference on Learning
                  Representations},
  year =     2021,
  month =     may,
  url =         {https://openreview.net/forum?id=gZ9hCDWe6ke},
}

@inproceedings{Zhu:2024:APC,
  author =     {Zhu, Haiyun and Liu, Jiaqi and Li, Xianxu and Huang,
                  Zhiqin and Zhang, Yong},
  title =     {A Power Consumption Measurement Method for Large
                  {AI}-based Intelligent Computing Servers},
  year =     2024,
  isbn =     9798400716485,
  publisher_ =     {Association for Computing Machinery},
  address_ =     {New York, NY, USA},
  doi =         {10.1145/3653081.3653107},
  booktitle =     {Proceedings of the 2023 5th International Conference
                  on Internet of Things, Automation and Artificial
                  Intelligence},
  pages =     {150--155},
  numpages =     6,
  location =     {Nanchang, China},
  series =     {IoTAAI '23},
}

@inproceedings{Ziming:2020:HHR,
  author =     {Liu, Ziming and Gao, Guangyu and Sun, Lin and Fang,
                  Zhiyuan},
  booktitle =     {2021 IEEE International Conference on Multimedia and
                  Expo (ICME)},
  title =     {{HRDNet}: High-Resolution Detection Network for
                  Small Objects},
  year =     2021,
  month =     jul,
  pages =     {1--6},
  doi =         {10.1109/ICME51207.2021.9428241}
}

@inproceedings{Alibeigi:2023:ZOD,
  title={Zenseact {O}pen {D}ataset: A large-scale and diverse multimodal dataset for autonomous driving},
  author={Alibeigi, Mina and Ljungbergh, William and Tonderski, Adam and Hess, Georg and Lilja, Adam and Lindstrom, Carl and Motorniuk, Daria and Fu, Junsheng and Widahl, Jenny and Petersson, Christoffer},
  booktitle={Proceedings of the IEEE/CVF International Conference on Computer Vision},
  year=2023,
  month= oct,
  pages={20121--20131},
  doi={10.1109/ICCV51070.2023.01846}
}


\end{document}